\documentclass[letterpaper, 10pt, conference]{ieeeconf}
\IEEEoverridecommandlockouts  % needed to use the \thanks command
\usepackage{algorithm}
\usepackage{amsmath}
\usepackage{amssymb}
\usepackage{bm}
\usepackage{color}
\usepackage{graphicx}
\usepackage{hyperref}

\newcommand{\dd}[1]{\ensuremath {{\rm d}{#1}}}
\newcommand{\defeq}{\stackrel{\mathrm{def}}{=}}

\newcommand{\tr}{\mathrm{tr}}

\def\bfcd{\dot{\bfc}}
\def\bfpd{\dot{\bfp}}

\def\bfrd{\dot{\bfr}}
\def\bfxid{\dot{\bfxi}}
\def\omegad{\dot{\omega}}

\def\sd{\dot{s}}
\def\xd{\dot{x}}

\def\bfcdd{\ddot{\bfc}}
\def\bfpdd{\ddot{\bfp}}

\def\bfrdd{\ddot{\bfr}}

\def\sdd{\ddot{s}}

\newcommand{\bfxi}{\boldsymbol{\xi}}

\newcommand{\bfpi}{\boldsymbol{\pi}}

\newcommand{\bfb}{\ensuremath {\bm{b}}}
\newcommand{\bfc}{\ensuremath {\bm{c}}}

\newcommand{\bfe}{\ensuremath {\bm{e}}}

\newcommand{\bfg}{\ensuremath {\bm{g}}}

\newcommand{\bfn}{\ensuremath {\bm{n}}}

\newcommand{\bfp}{\ensuremath {\bm{p}}}

\newcommand{\bfr}{\ensuremath {\bm{r}}}

\newcommand{\bfA}{\mathbf{A}}

\usepackage{empheq}

\definecolor{bkgnd-color}{cmyk}{0.0, 0.04, 0.06, 0}
\definecolor{title-color}{cmyk}{0.1, 0.0, 0.0, 0}

\newsavebox{\mysaveboxM} % M for math
\newsavebox{\mysaveboxT} % T for text
\newcommand*\Garybox[2][Example]{%
    \sbox{\mysaveboxM}{#2}%
    \sbox{\mysaveboxT}{\fcolorbox{black}{title-color}{#1}}%
    \sbox{\mysaveboxM}{%
        \parbox[b][\ht\mysaveboxM+.5\ht\mysaveboxT+.5\dp\mysaveboxT][b]{%
        \wd\mysaveboxM}{#2}%
    }%
    \sbox{\mysaveboxM}{%
        \fcolorbox{black}{bkgnd-color}{%
            \makebox[\linewidth-5em]{\usebox{\mysaveboxM}}%
        }%
    }%
    \usebox{\mysaveboxM}%
    \makebox[0pt][r]{%
        \makebox[\wd\mysaveboxM][c]{%
            \raisebox{\ht\mysaveboxM-0.5\ht\mysaveboxT +0.5\dp\mysaveboxT-0.5\fboxrule}{\usebox{\mysaveboxT}}%
        }%
    }%
}

\def\bfcbard{\dot{\bar{\bfc}}}
\def\bfcbar{\bar{\bfc}}
\def\bfcf{\bfc_\fsubscript}
\newcommand{\bfvarphi}{\boldsymbol{\varphi}}
\def\bfci{\bfc_\isubscript}
\def\bfcdi{\bfcd_\isubscript}
\def\bfpi{\bfp_\isubscript}
\def\bfpdi{\dot{\bfp}_\isubscript}
\def\bfri{\bfr_\isubscript}
\def\bfrf{\bfr_\fsubscript}
\def\fsubscript{\mathrm{f}}
\def\isubscript{\mathrm{i}}
\def\lambdai{\lambda_\isubscript}
\def\lambdamax{\lambda_\text{max}}
\def\lambdamin{\lambda_\text{min}}
\def\omegaimax{\omega_{\isubscript,\text{max}}}
\def\omegaimin{\omega_{\isubscript,\text{min}}}
\def\omegai{\omega_{\isubscript}}

\def\xdi{\dot{x}_\isubscript}
\def\zdi{\dot{z}_\isubscript}
\def\zf{z_\fsubscript}
\def\zi{z_\isubscript}
\floatname{algorithm}{Optimization Problem}
\title{\LARGE \bf
    Balance control using both ZMP and COM height variations: \\
    A convex boundedness approach
}

\author{St\'ephane Caron and Bastien Mallein%
    \thanks{This work is supported in part by the H2020 EU project COMANOID
    \url{http://www.comanoid.eu/}, RIA No 645097.}%
    \thanks{S. Caron is with the Montpellier Laboratory of Informatics,
    Robotics and Microelectronics, CNRS--University of Montpellier,
    Montpellier, France.}%
    \thanks{B. Mallein is with the Laboratoire Analyse, G\'{e}om\'{e}trie et
    Applications, CNRS--Paris 13 University, Villetaneuse, France.}%
    \thanks{Corresponding author: {\tt\footnotesize
    stephane.caron@lirmm.fr}}%
}

\begin{document}

\maketitle
\thispagestyle{empty}
\pagestyle{empty}

\begin{abstract}
    Developments for 3D control of the center of mass (CoM) of biped robots are
    currently located in two local minima:~on the one hand, methods that allow
    CoM height variations but only work in the 2D sagittal plane; on the other
    hand, nonconvex direct transcriptions of centroidal dynamics that are
    delicate to handle. This paper presents an alternative that controls the
    CoM in 3D via an indirect transcription that is both low-dimensional and
    solvable fast enough for real-time control. The key to this development is
    the notion of boundedness condition, which quantifies the capturability of
    3D CoM trajectories.
\end{abstract}

\section{Introduction}

Three-dimensional control of the center of mass follows in the wake of major
achievements obtained in 2D locomotion with the linear inverted pendulum
mode~(LIPM)~\cite{kajita20013d}. The core idea of the LIPM was to keep the CoM
in a plane, which made the model tractable and paved the way for key
discoveries, including the capture point~\cite{pratt2006capture} and
capture-point based feedback control~\cite{sugihara2009standing}, subsequently
applied in successful walking controllers~\cite{morisawa2012balance,
englsberger2015three}.

For a while, the ability to leverage vertical CoM motions seemed lost on the
way, but recent developments showed a regain of interest for this
capability~\cite{ramos2015generalizations, koolen2016humanoids,
gao2017increase}. All of them share a design choice that can be traced back to
the seminal work of Pratt and Drakunov~\cite{pratt2007derivation}: they
interpolate CoM trajectories in a 2D sagittal plane for the inverted pendulum
model (IPM) with fixed center of pressure. The key result
of~\cite{pratt2007derivation} is the conservation of the ``orbital energy'' of
a CoM path, a variational principle that was later translated into a predictive
controller in an equally inspirational study by Koolen \emph{et
al.}~\cite{koolen2016humanoids}. Ramos and
Hauser~\cite{ramos2015generalizations} also noticed that the capture point,
interpreted as \emph{point where to step}, was a function of the CoM path,
which they computed via a single shooting method. Interestingly, Hopkins
\emph{et al.} pointed out that vertical CoM motions is equivalent to turning
the constant $\omega$ of the LIPM into a time-varying function
$\omega(t)$~\cite{hopkins2014humanoid}. They brought to light a differential
equation that this function must satisfy, and used it to compute back
$\omega(t)$ from, once again, an \emph{a priori} CoM-height trajectory.

Different as they may seem, the variational and point-where-to-step approaches
are two instances of the same underlying concept: convergence of the system
towards a steady state requires that its divergent component of motion stays
bounded. Lanari \emph{et al.}~\cite{lanari2014boundedness, lanari2015iros}
showed how the condition for this to happen involves an integral over the
future trajectory of the system (one can see a similar integral in the orbital
energy), which they named the \emph{boundedness condition}.

The solution we explore in this study differs from existing approaches in that
it considers $\omega(t)$ \emph{per se} rather than as a result of CoM
trajectories. This change of perspective is made possible by the derivation of
the boundedness condition for the 3D IPM (Section~\ref{sec:boundedness}), which
is then cast into an optimization problem for 2D and 3D control of the CoM
(Sections~\ref{sec:2d-balance} and~\ref{sec:3d-balance}). The resulting
predictive controller is implemented and tested in Section~\ref{sec:simus}.

\section{Boundedness condition of the IPM}
\label{sec:boundedness}

The equation of motion of the inverted pendulum model is:
\begin{equation}
    \label{eq:ipm}
    \bfcdd(t) = \lambda(t) (\bfc(t) - \bfr(t)) + \bfg
\end{equation}
where $\bfc$ is the center of mass (CoM) of the robot, $\bfr$ is the center of
pressure (CoP) under its contacting foot, and $\bfg = -g\bfe_z$ is the gravity
vector. The quantity $\lambda$ has the unit of a stiffness. It must be positive
$\lambda \geq 0$ by unilaterality of contact, while the CoP $\bfr$ always
belongs to the contact area.

\subsection{Divergent component of motion}

To alleviate calculations, let us formulate the divergent component of
motion~\cite{takenaka2009iros} of this model as a velocity rather than a
point:\footnote{Consider the derivative of a product $u v$ compared to that of
a ratio $u / v$.}
\begin{equation}
    \label{eq:dcm}
    \bfxi(t) = \bfcd(t) - \bfrd(t) + \omega(t) (\bfc(t) - \bfr(t))
\end{equation}
where $\omega(t)$ is a solution to the differential
equation~\cite{hopkins2014humanoid}:
\begin{equation}
    \label{eq:omegad}
    \omegad = \omega^2 - \lambda
\end{equation}
The quantity $\omega$ has the unit of a damping. The interest of its
differential equation appears when differentiating the divergent component of
motion:
\begin{equation}
    \bfxid = (\lambda + \omegad) (\bfc - \bfr) + \omega (\bfcd - \bfrd) + \bfg
    - \bfrdd = \omega \bfxi + \bfg - \bfrdd
\end{equation}
The solution to this first-order differential equation is:
\begin{equation}
    \label{eq:xi-solution}
    \bfxi(t) = \left(\bfxi(0) + \int_0^t e^{-\Omega(\tau)} (\bfg -
    \bfrdd(\tau))
    \dd{\tau}\right) e^{\Omega(t)}
\end{equation}
where $\Omega$ is the antiderivative of $\omega$ such that $\Omega(0)=0$. In
the LIPM where the damping $\omega$ is a constant, $\Omega(t) = \omega t$ and
this equation is equivalent to the well-known capture point dynamics~\cite{takenaka2009iros}. 

To be viable~\cite{wieber2016modeling}, the trajectory of the system must be
bounded, which implies that the above expression does not diverge despite its
exponential factor. This necessary condition for viability is known as the
\emph{boundedness condition}~\cite{lanari2014boundedness, lanari2015iros}, and
is written here:
\begin{equation}
    \label{eq:boundedness1}
    \int_0^\infty (\bfrdd(t) - \bfg) e^{-\Omega(t)} \dd{t} = \bfxi(0)
\end{equation}
Note how this requirement involves both the CoP trajectory $\bfr(t)$ and
stiffness trajectory $\lambda(t)$: the former is integrated through
$\bfrdd(t)$, while the latter is embedded in $\Omega(t)$.

\subsection{Decoupling of the boundedness constraint}

\begin{figure}[t]
    \centering
    \includegraphics[width=0.8\columnwidth]{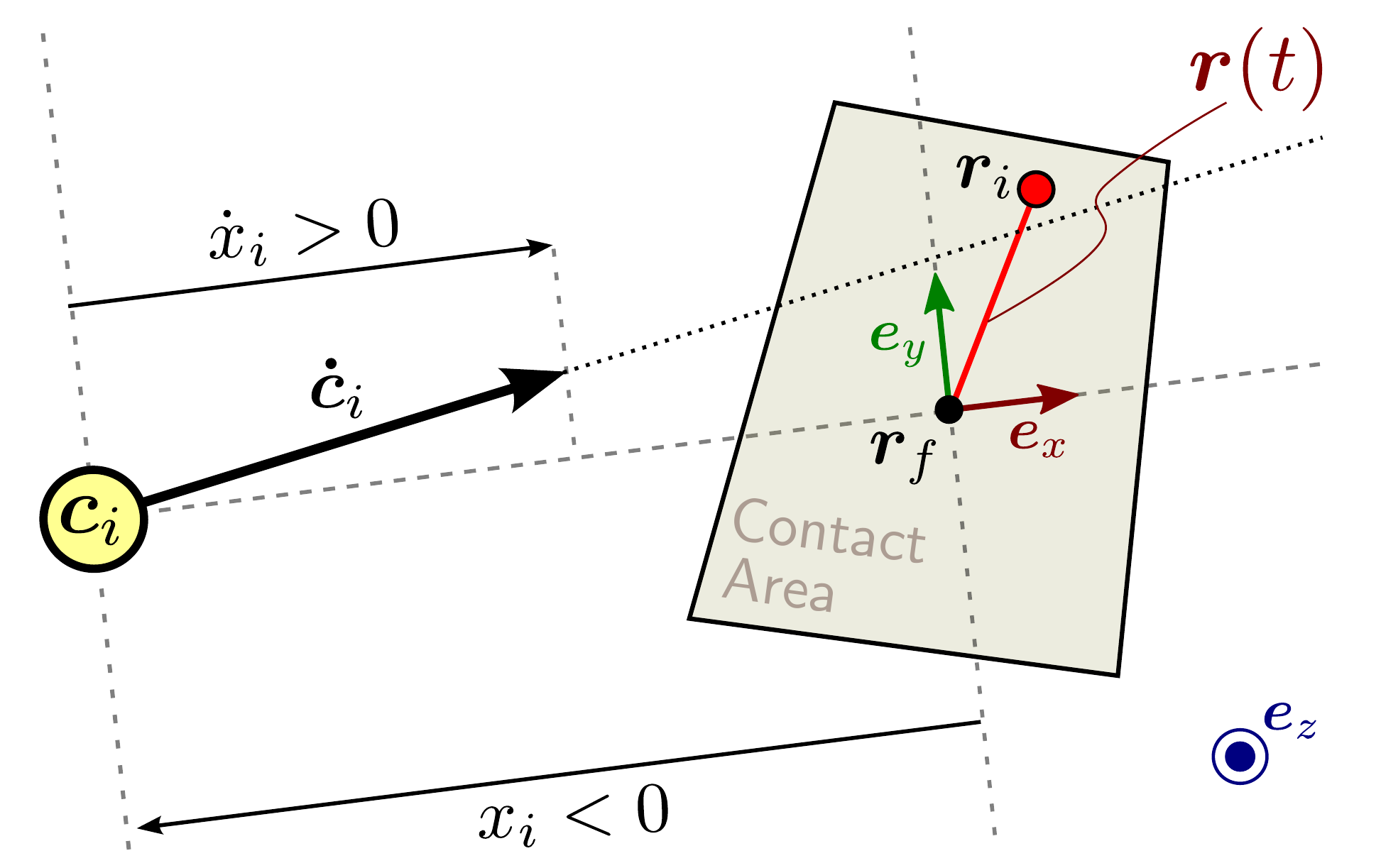}
    \caption{
        \textbf{Definition of the inertial frame $(\bfe_x, \bfe_y, \bfe_z)$.}
        The frame is rooted at the stationary CoP location $\bfr_f$. The
        instantaneous CoM location is denoted by $\bfci$. Both vectors
        $\bfe_x$ and $\bfe_y$ lie in the horizontal plane, and $\bfe_x$ is
        aligned with the (horizontal projection of the) desired direction of
        motion $\bfrf - \bfci$.
    }
    \label{fig:inertial-frame}
\end{figure}

Let us express all coordinates in the inertial frame depicted in
Figure~\ref{fig:inertial-frame}. In what follows, we will use the subscript
$\isubscript$ (``initial'') to indicate values at $t=0$, and $\fsubscript$
(``final'') for stationary values obtained as $t \to \infty$. The origin of the
inertial frame is taken at the stationary CoP position $\bfrf = \bm{0}$ where
we want the robot to stop. The stationary CoM position is then $\bfcf = \zf
\bfe_z$, with $\zf$ the desired stationary CoM height. The two vectors $\bfe_x$
and $\bfe_y$ are horizontal (\emph{i.e.} orthogonal to gravity), and $\bfe_x$
is chosen aligned with the horizontal projection of $\bfrf - \bfci$.

In reality, the center of pressure always lies on the contact surface between
the robot and its environment. Therefore, CoPs $\bfr$ output by our controller
must satisfy the \emph{feasibility} condition of lying on the contact surface.
Denoting by $\bfn$ the contact normal, this means that $\bfr$ and all its
derivatives are orthogonal to $\bfn$; in particular, $\bfrdd(t) \cdot \bfn = 0$
at all times. Taking the dot product of Equation~\eqref{eq:boundedness1} with
$\bfn$ then yields:
\begin{align}
    \int_0^\infty e^{-\Omega(t)} \dd{t} 
    & = \frac{\dot{\bar{z}}_\isubscript + \omegai \bar{z}_\isubscript}{g}
\end{align}
where $\bar{z} \defeq {(\bfc \cdot \bfn)} / {(\bfe_z \cdot \bfn)}$ is the
height of the vertical projection of $\bfc$ onto the contact surface. 

Let us now define $\bfp(t) \defeq \begin{bmatrix} r_x(t) & r_y(t)
\end{bmatrix}$ and $\bfcbar_\isubscript \defeq \begin{bmatrix} x_\isubscript &
y_\isubscript \end{bmatrix}$ the horizontal projections of the CoP and CoM,
respectively. The horizontal projection of Equation~\eqref{eq:boundedness1} is:
\begin{align}
    \int_0^\infty \bfpdd(t) e^{-\Omega(t)} \dd{t} = (\bfcbard_\isubscript - \bfpd_\isubscript) + \omegai (\bfcbar_\isubscript - \bfp_\isubscript)
\end{align}
A double integration by parts of the left hand side of this equation yields:
\begin{equation*}
    \int_0^\infty \bfpdd(t) e^{-\Omega(t)} \dd{t}
    = \int_0^\infty \bfp(t) \lambda(t) e^{-\Omega(t)} \dd{t}
    - (\bfpd_\isubscript + \omegai \bfp_\isubscript) 
\end{equation*}
Combining these last two equations, we obtain a temporal formulation of the
boundedness condition with separate gravity and CoP components:
\begin{empheq}[box={\Garybox[Temporal Boundedness Condition]}]{align}
    \int_0^\infty e^{-\Omega(t)} \dd{t}
    & = \frac{\dot{\bar{z}}_\isubscript + \omegai \bar{z}_\isubscript}{g}
    \label{eq:bc-time-1}
    \\
    \int_0^\infty \bfp(t) \lambda(t) e^{-\Omega(t)} \dd{t}
    & = \bfcbard_\isubscript + \omegai \bfcbar_\isubscript 
    \label{eq:bc-time-2}
\end{empheq}
Equations~\eqref{eq:bc-time-1}--\eqref{eq:bc-time-2} are equivalent
to~\eqref{eq:boundedness1} as $(\bfe_x, \bfe_y, \bfn)$ is a (non-orthogonal)
basis of the 3D Euclidean space.

In the familiar setting of the LIPM, the first equation boils
down to:
\begin{equation}
    \int_0^\infty e^{-\omega t} \dd{t} = \frac{1}{\omega} = \frac{0 + \omega
    z}{g}
    \Longleftrightarrow
    \omega = \sqrt{\frac{g}{z}}
\end{equation}
Meanwhile, taking $\bfp=0$ in the second one implies that the traditional LIPM
capture point is located at $\bfr_f$. This observation illustrates how the
boundedness condition~\eqref{eq:bc-time-1}--\eqref{eq:bc-time-2} encompasses
the well-known model.

\subsection{Change of variable}

Define the adimensional quantity $s(t) = e^{-\Omega(t)}$. This variable ranges
from $1$ when $t=0$ to $0$ as $t \to \infty$, and its time derivatives are:
\begin{align}
\sd(t) & = - \omega(t) s(t) \label{eq:sd} \\
\sdd(t) & = \lambda(t) s(t) \label{eq:sdd}
\end{align}
Thanks to the bijective mapping between $s$ and $t$, we can characterize
$\omega$ and $\lambda$ indistinctly as functions of $s$ or $t$. Let us choose
the former, and denote by $(\cdot)'$ derivatives with respect to $s$. For
instance, $\omega(t) = \omega(s(t))$ implies that:
\begin{align}
    \frac{\dd{\omega}}{\dd{t}}(t) 
    = \frac{\dd{s}}{\dd{t}}(t) \frac{\dd{\omega}}{\dd{s}}(s(t))
    \Longrightarrow 
    \omegad = \sd \omega' = - s \omega \omega'
\end{align}
where we eventually omit arguments $\square(t)$ and $\square(s(t))$ for
brevity. Applying the chain rule similarly yields:
\begin{align}
    \dd{t} & = \frac{\dd{s}}{\sd} = -\frac{\dd{s}}{\omega(s) s}
    \\
    \lambda & = \omega^2 - \omegad = \omega (\omega + s \omega') = \omega (s
    \omega)'
    \label{eq:lambda-s}
\end{align}
We can therefore formulate the boundedness
condition~\eqref{eq:bc-time-1}--\eqref{eq:bc-time-2} in terms of the new
variable $s$ as:
\begin{empheq}[box={\Garybox[Timeless Boundedness Condition]}]{align}
     \int_0^1 \frac{1}{\omega(s)} \dd{s} & = \frac{\dot{\bar{z}}_\isubscript + \omegai
     \bar{z}_\isubscript}{g}
     \label{eq:bc-s-1}
     \\
     \int_0^1 \bfp(s) (s \omega)' \dd{s} & = \bfcbard_\isubscript + \omegai \bfcbar_\isubscript 
     \label{eq:bc-s-2}
\end{empheq}
With this change of variable, infinite integrals over time solutions
$\omega(t), \bfp(t)$ are turned into finite integrals over functions
$\omega(s), \bfp(s)$ of the new variable $s$. This approach bears a close
resemblance to the recently proposed concept of \emph{spatial
quantization}~\cite{kajita2017humanoids}.

\section{Sagittal 2D Balance with Fixed CoP}
\label{sec:2d-balance}

Let us consider first the case of 2D stabilization in the plane $(\bfe_x,
\bfe_z)$ with a fixed CoP. Since $r_x=0$, the boundedness condition becomes:
\begin{align}
    \omegai & = -\frac{\xdi}{x_\isubscript} \label{eq:2d-1} \\
    \zdi + \omegai \zi & = g \int_0^1 \frac{1}{\omega(s)} \dd{s} \label{eq:2d-2}
\end{align}
Equation~\eqref{eq:2d-1} implies via \eqref{eq:xi-solution} that $\xd = -
\omega x$, which integrates to $x(t) = x_\isubscript e^{-\Omega(t)} =
x_\isubscript s(t)$. We see how, in 2D balance control, the coordinates $x$ and
$s$ are proportional. Interpolation over $x$~\cite{koolen2016humanoids,
pratt2007derivation} is therefore equivalent to interpolation over $s$, and one
can recognize in~\eqref{eq:2d-2} the same principle as the conservation of
orbital energy~\cite{pratt2007derivation}. The benefit of using $s$ rather than
$x$ will appear when we move to 3D control.

\subsection{Viability condition}

The viability condition $z_\text{crit} \geq 0$ derived
in~\cite{koolen2016humanoids} for $\lambda \geq 0$ can be seen as a consequence
of~\eqref{eq:2d-2}. Indeed, the lower-bounding profile $\underline{\lambda}=0$
corresponds to $(s \underline{\omega})'=0$, that is to say, $s
\underline{\omega} = \omegai$. It follows that:
\begin{gather}
    \int_0^1 \frac{\dd{s}}{\omega(s)} \geq \int_0^1
    \frac{\dd{s}}{\underline{\omega}(s)} = \int_0^1 \frac{s}{\omegai}  \dd{s} =
    \frac{1}{2\omegai} \\
    z_\text{crit} = \zi + \frac{\zdi}{\omegai} - \frac{g}{2 \omegai^2}
    = \frac{g}{\omegai}\left[\int_0^1 \frac{\dd{s}}{\omega(s)} - \frac{1}{2
    \omegai}\right] \geq 0
\end{gather}
In practice, biped robots cannot exert arbitrary large contact forces, and tend
to break contact when ground pressure becomes too low. To reflect this, we will
thereafter consider the stricter feasibility condition $\lambda \in
[\lambdamin, \lambdamax]$. The corresponding viability condition can be derived
in a similar fashion by considering lower- and upper-bounding
profiles:\footnote{
    A more detailed derivation is provided in the supplementary material:
    \url{https://scaron.info/files/icra-2018/supmat.pdf}}
\begin{align}
    \zdi + \omegai \zi & \leq \frac{g}{\lambdamax}\left[\omegai +
    \sqrt{\frac{\lambdamax - \lambdamin}{\lambdamin}}\sqrt{\lambdamax-\omegai^2}\right]
    \\
    \zdi + \omegai \zi & \geq \frac{g}{\lambdamin} \left[\omegai -
    \sqrt{\frac{\lambdamax - \lambdamin}{\lambdamax}} \sqrt{\omegai^2 -
    \lambdamin}\right]
\end{align}

\subsection{Computing feasible \texorpdfstring{$\omega$}{omega} solutions}

Let us partition the interval $[0, 1]$ into $N$ fixed segments $0 = s_0 < s_1
< \ldots < s_N = 1$, for instance $s_j = j / N$. We compute solutions
to the boundedness condition where $\lambda(s)$ is piecewise constant over this
subdivision, that is, $\forall s \in [s_j, s_{j+1}], \lambda(s) = \lambda_j$.
Define:
\begin{align}
    \varphi(s) & \defeq s^2 \omega^2 &
    \Delta_j & \defeq s_{j+1}^2 - s_j^2
\end{align}
Remarking that $\varphi' = 2 s \lambda$ from Equation~\eqref{eq:lambda-s}, we can
directly compute $\varphi(s)$ for $s \in [s_j, s_{j+1}]$ as:
\begin{equation}
    \label{eq:varphi-s}
    \varphi(s) = \sum_{k=0}^{j-1} \lambda_k \Delta_k + \lambda_j (s^2 - s_j^2)
\end{equation}
In what follows, we use the shorthand $\varphi_j = \varphi(s_j)$.

We can now calculate the right-hand side of the boundedness
condition~\eqref{eq:2d-2}:
\begin{align}
    \int_0^1 \frac{\dd{s}}{\omega(s)}
    & = \sum_{j=0}^{N-1} \int_{s_j}^{s_{j+1}} \frac{s \dd{s}}{\sqrt{\varphi_j +
    \lambda_j (s^2 - s_j^2)}}
    \\
    & = \sum_{j=0}^{N-1} \int_{0}^{\Delta_j} \frac{\dd{v}}{2 \sqrt{\varphi_j +
    \lambda_j v}}
    \\
    & = \sum_{j=0}^{N-1} \frac{1}{\lambda_j} \left[\sqrt{\varphi_j + \lambda_j \Delta_j} -
    \sqrt{\varphi_j}\right]
    \\
    & = \sum_{j=0}^{N-1} \frac{\Delta_j}{\sqrt{\varphi_{j+1}} + \sqrt{\varphi_j}}
    \label{eq:conv-obj}
\end{align}
Note that the latter expression is convex in the variables $\varphi_1, \ldots,
\varphi_N$ (by definition $\varphi_0 = 0$), as shown in
Appendix~\ref{convexity-proof}.

Besides boundedness, solutions should enforce three conditions:
\begin{itemize}
    \item \textit{Feasibility:} $\lambdamin \leq \lambda \leq \lambdamax$,
        expressed linearly in terms of $\varphi$ as $\Delta_j \lambdamin \leq
        \varphi_{j+1} - \varphi_j \leq \Delta_j \lambdamax$;
    \item \textit{Initial state:} $\omega$ should be equal to $\omegai$ at the
        initial index $s=1$, that is to say, $\varphi_N = \omegai^2$;
    \item \textit{Stationary state:} a stationary COM height $\zf$ can also be
        specified via $\varphi_1 = \Delta_0 g / \zf$, or similarly a range of
        heights $z_\text{min} \leq \zf \leq z_\text{max}$ used to approximate
        kinematic reachability.
\end{itemize}
Wrapping all four conditions together and adding a regularizing cost function
over variations of $\lambda$, we obtain Optimization Problem~\ref{cvx-prob-2d}.
This problem is ``almost'' a quadratic program: it has a quadratic cost
function and linear constraints, except for Equation~\eqref{eq:conv-cons-2d}
which is a one-dimensional nonlinear equality constraint.

\subsection{Model predictive control for 2D balance}

Solving Problem~\ref{cvx-prob-2d} can be done fast enough for the control loop,
on the scale of 1--3~ms using a general-purpose nonlinear solver (see
Section~\ref{sec:simus}). At each control cycle, we
compute the optimum $\varphi_1^*, \ldots, \varphi_N^*$ of the problem and extract its
initial stiffness via:
\begin{equation}
    \lambdai^* = \frac{\omegai^2 - \varphi_{N-1}^*}{\Delta_{N-1}}
\end{equation}
This value is then sent as reference to the lower-level leg and attitude
controllers until the next control cycle. In the standard model-predictive
fashion, the rest of the optimal trajectory is discarded. As a matter of fact,
the trajectories $\omega(s)$ and $\lambda(s)$ are never explicitly computed,
let alone their time counterparts. The operation is possible (see
Appendix~\ref{computing-time-traj}) but not necessary for control. Compared to
the controller from~\cite{koolen2016humanoids}, this solution enforces
feasibility constraints $\lambdamin \leq \lambda \leq \lambdamax$ \emph{a
priori}, as opposed to an \emph{a posteriori} clipping that may cause
free-falling phases in the output trajectory.

\begin{algorithm}[t]
    \caption{\hfill 2D Balance Control}
    \label{cvx-prob-2d}
    \begin{align}
        \underset{\varphi_1, \ldots, \varphi_N}{\text{minimize }}
        & \sum_{j=1}^{N-1} \left[ 
        \frac{\varphi_{j+1}-\varphi_j}{\Delta_j} - \frac{\varphi_j - \varphi_{j-1}}
        {\Delta_{j-1}} \right]^2
        \label{eq:cvx-cost}
        \\
        \text{subject to }
        &
        \sum_{j=0}^{N-1} \frac{\Delta_j}{\sqrt{\varphi_{j+1}} + \sqrt{\varphi_j}} 
        = \frac{\dot{\bar{z}}_\isubscript + \omegai \bar{z}_\isubscript}{g}
        \label{eq:conv-cons-2d} \\
        &
        \varphi_N = \omegai^2 \label{eq:omega-i-2d} \\
        &
        \forall j,\ \lambdamin \Delta_j \leq \varphi_{j+1} - \varphi_j \leq
        \lambdamax \Delta_j \label{eq:cvx-lambda-ineq} \\
        &
        \varphi_1 = \Delta_0 g / \zf \label{eq:cvx-stat}
    \end{align}
\end{algorithm}

\section{3D Balance}
\label{sec:3d-balance}

In 3D balance where the initial velocity $\bfcdi$ is not necessarily coplanar
with the stationary state, the CoP cannot be fixed and the initial damping
$\omegai$ is not determined by Equation~\eqref{eq:2d-1} any more. The latter is
replaced by a more general condition over CoP trajectories:
\begin{equation}
    \int_0^1 \bfp(s) (s \omega)' \dd{s} = \bfcbard_{i} + \omegai \bfcbar_\isubscript
    \label{eq:bc-recap}
\end{equation}
Subject to the feasibility conditions:
\begin{equation}
    \label{eq:cop-feas}
    \bfA \bfp \leq \bfb
\end{equation}
where the matrix $\bfA$ and vector $\bfb$ form the halfspace-representation of
the contact polygon. This polygon is readily computed from contact geometry.
For example, a rectangular contact $|(\bfp \cdot \bfe_w)| \leq W$, $|(\bfp
\cdot \bfe_h)| \leq H$ has four inequalities:
\begin{equation}
    \pm \begin{bmatrix}
        (\bfe_x \cdot \bfe_w) & (\bfe_y \cdot \bfe_w) \\
        (\bfe_x \cdot \bfe_h) & (\bfe_y \cdot \bfe_h)
    \end{bmatrix}
    \bfp  
    \leq \begin{bmatrix} W \\ H \end{bmatrix}
\end{equation}

\subsection{Computing feasible CoP solutions}

The structure of Equation~\eqref{eq:bc-recap} suggests a particular
solution:~let us take $\bfp(s) = \bfpi f(s \omega)$, where:
\begin{itemize}
    \item $f(\omegai) = 1$: initially, the CoP is located at $\bfri$;
    \item $f(0) = 0$: eventually, the CoP is located at $\bfrf$;
    \item \emph{$f$ is increasing:} we exclude solutions where the CoP goes
        back and forth, that we deem suboptimal;
    \item \emph{$f$ is integrable:} let $F$ denote its antiderivative such that
        $F(0) = 0$. It is positive by monotonicity of $f$.
\end{itemize}
With this choice, the boundedness condition~\eqref{eq:bc-recap} becomes
\begin{equation}
    \label{eq:new-cop-bc}
    \bfpi = \frac{\bfcbard_\isubscript + \omegai \bfcbar_\isubscript}{F(\omegai)}
\end{equation}
By monotonicity of $f$, the CoP trajectory is feasible if and only if its
initial position $\bfpi$ satisfies the feasibility
condition~\eqref{eq:cop-feas}. Considering~\eqref{eq:new-cop-bc}, this can be
written:
\begin{equation}
    \label{eq:F-omega-geq}
    \bfb F(\omegai) - (\bfA \bfcbar_\isubscript) \omegai \geq \bfA \bfcbard_\isubscript
\end{equation}
Compared to the previous fixed-CoP setting where $\omegai$ was fully
determined, we see how the relaxed polygonal constraint $\bfA \bfp \leq \bfb$
now frees a range of possible choices for $\omegai$. At this stage, one can
explore different CoP strategies via the choice of a (preferably convex)
function $F$. Let us focus on the example of a power law, \emph{i.e.} for some
$k > 1$:
\begin{equation}
    f(s \omega) = \left(s \frac{\omega}{\omegai}\right)^{k-1}
    \Longrightarrow F(\omegai) = \frac{\omegai}{k}
    \label{eq:f-and-F}
\end{equation}
Equation~\eqref{eq:F-omega-geq} simplifies to:
\begin{equation}
    \left( \frac{\bfb}{k} - \bfA \bfcbar_\isubscript\right) \omegai \geq \bfA \bfcbard_\isubscript
\end{equation}
Each line of this vector inequality provides a lower or upper bound on
$\omegai$, depending on the sign of the factor in front of it. These
inequalities can then be summed up as $\omegaimin \leq \omegai \leq
\omegaimax$. Note that this computation is only carried out once from the
initial state, \emph{i.e.}~it is not part of the following numerical
optimization.

\begin{algorithm}[t]
    \caption{\hfill3D Balance Control}
    \label{cvx-prob-3d}
    \begin{align}
        \underset{\varphi_1, \ldots, \varphi_N}{\text{minimize }}
        & \sum_{j=1}^{N-1} \left[ 
        \frac{\varphi_{j+1}-\varphi_j}{\Delta_j} - \frac{\varphi_j - \varphi_{j-1}}
        {\Delta_{j-1}} \right]^2
        \tag{\ref{eq:cvx-cost}}
        \\
        \text{subject to }
        &
        \sum_{j=0}^{N-1} \frac{\Delta_j}{\sqrt{\varphi_{j+1}} + \sqrt{\varphi_j}} 
        - \frac{\bar{z}_\isubscript}{g} \sqrt{\varphi_N} = \frac{\dot{\bar{z}}_\isubscript}{g} \\
        &
        \omegaimin^2 \leq \varphi_N \leq \omegaimax^2 \label{eq:omega-i-3d} \\
        &
        \forall j,\ \lambdamin \Delta_j \leq \varphi_{j+1} - \varphi_j \leq
        \lambdamax \Delta_j \tag{\ref{eq:cvx-lambda-ineq}} \\
        &
        \varphi_1 = \Delta_0 g / \zf \tag{\ref{eq:cvx-stat}}
    \end{align}
\end{algorithm}

By definition of $\varphi$, $\omegai$ is equal to $\sqrt{\varphi_N}$, so that bounds
on $\omega$ are mapped directly into the optimization as $\omegaimin^2 \leq
\varphi_N \leq \omegaimax^2$. However, previously $\omegai$ also appeared in
the right-hand side of the nonlinear equality
constraint~\eqref{eq:conv-cons-2d}. In 3D, this constraint becomes:
\begin{equation}
    \sum_{j=0}^{N-1} \frac{\Delta_j}{\sqrt{\varphi_{j+1}} + \sqrt{\varphi_j}} 
    - \frac{\bar{z}_\isubscript}{g} \sqrt{\varphi_N} = \frac{\dot{\bar{z}}_\isubscript}{g}
\end{equation}
Both the function $x \mapsto -\sqrt{x}$ and the expression from
Equation~\eqref{eq:conv-obj} are convex, therefore this new expression is
convex as well. Wrapping up these developments, we obtain Optimization
Problem~\ref{cvx-prob-3d}.

\subsection{Model predictive control for 3D balance}

Our pipeline for 3D balance is the same as in Section~\ref{sec:2d-balance}.
Once the optimal solution $\varphi_1^*, \ldots, \varphi_N^*$ of
Problem~\ref{cvx-prob-3d} is found, stiffness and CoP are extracted as:
\begin{align}
    \omegai^* & = \sqrt{\varphi_N^*} &
    \lambdai^* & = \frac{\varphi_N^* - \varphi_{N-1}^*}{\Delta_{N-1}} &
    \bfpi^* & = \frac{\bfcbard_\isubscript + \bfcbar_\isubscript \omegai^*}{F(\omegai^*)}
\end{align}
In the case of the power law~\eqref{eq:f-and-F}, the CoP solution becomes:
\begin{equation}
    \bfp^* = k \left[\bfpi + \frac{\bfpdi}{\omegai^*}\right]
    \label{eq:p-with-k}
\end{equation}
Interestingly, we recognize here the expression of the capture-point feedback
control law~\cite{sugihara2009standing, englsberger2015three,
takenaka2009iros}, under the usual requirement that $k > 1$, even though we are
in the context of 3D balance where $\lambda$ and $\omega$ are time-varying.

\begin{figure}[t]
    \centering
    \includegraphics[width=0.98\columnwidth]{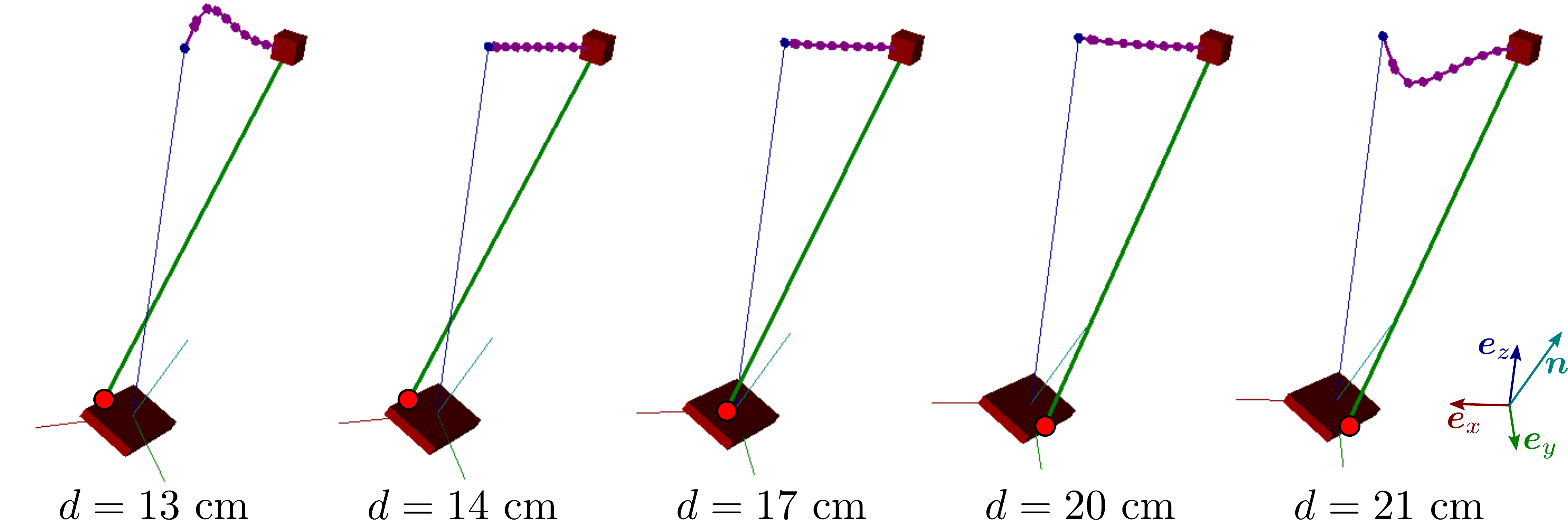}
    \caption{
        \textbf{Saturation behavior of the 3D balance controller.} The
        horizontal distance from CoM to contact is denoted by $d$, $\bfn$ is
        the contact normal and red discs indicate the initial CoP. The
        controller keeps the CoM trajectory as close as possible to a LIPM via
        CoP variations. When this is not enough, height variations are resorted
        to for additional braking.
    }
    \label{fig:saturation}
\end{figure}

Problem~\ref{cvx-prob-3d} is solved as fast as its 2D counterpart, on the scale
of 1--3~ms, but it is able to cope with both sagittal and lateral velocity
compensation. Its solutions tend to be flatter thanks to a wider range of CoP
positions. As a matter of fact, by trying to keep the stiffness $\lambda$ as
constant as possible, Problem~\ref{cvx-prob-3d} generates a hierarchical
strategy: CoP variations are used first; then, if need be, CoM height
variations are resorted to. This behavior is depicted in
Figure~\ref{fig:saturation}.

\subsection{Discussion}

CoM trajectory generation using 6D contacts is a notoriously nonconvex problem
due to angular momentum. Even when a linear model is used and angular momentum
is kept constant~\cite{englsberger2015three}, nonconvexity lingers in
feasibility inequality constraints~\cite{caron2016humanoids}. Nonlinear optimal
control has been explored on \emph{direct transcriptions} of this problem using
\emph{e.g.}~multiple shooting~\cite{carpentier2016versatile}; however, our
experience in~\cite{caron2017iros} met with frequent solver failures caused by
local optimum switches, which required dedicated countermeasures. To avoid such
switches, Ponton \emph{et al.}~\cite{ponton2016convex} proposed a convex
relaxation of the direct transcription of centroidal dynamics by bounding the
convex and concave parts of the angular momentum.

Compared to these previous works, the study we propose here relies on an
indirect transcription, \emph{i.e.} optimization variables are neither the
CoM position nor its derivatives. To reduce further the dimension of the
problem, we made the CoP trajectory a \emph{consequence} $\bfp(s) = \bfpi f(s
\omega)$ of the damping profile $\omega$, as opposed to \emph{e.g.}~a shooting
method where $\bfp$ and $\omega$ would be optimized jointly. Choosing a linear
CoP trajectory also impacts the existence of feasible solutions, for instance
when the feedback gain $k$ is too large. An effect that is by the way also
present in all capture-point feedback controllers~\cite{sugihara2009standing,
morisawa2012balance, englsberger2015three, hopkins2014humanoid,
pajon2017walking}.

Compared to \emph{e.g.}~\cite{caron2016humanoids, carpentier2016versatile,
caron2017iros}, we did not explicitly model frictional constraints in the
present study. We observe however, as noted in~\cite{koolen2016humanoids}, that
the friction force is maximal at the beginning of the balancing trajectory. It
can therefore be constrained (if needed) within the presented framework via
additional instantaneous CoP inequalities~\eqref{eq:cop-feas}.

\section{Stepping experiments}
\label{sec:simus}

We implemented both balance controllers\footnote{
    \url{https://github.com/stephane-caron/3d-balance}
} and evaluated them in
\emph{pymanoid}\footnote{
    \url{https://github.com/stephane-caron/pymanoid}
}, an extension of OpenRAVE for humanoid robotics. Optimization problems were
formulated with a spatial discretization $s_j = j / N$ with $N = 10$. They were
subsequently solved using the IPOPT solver,\footnote{
    \url{https://projects.coin-or.org/Ipopt}
} with Jacobians and Hessians computed by automatic differentiation via
CasADi.\footnote{
    \url{http://casadi.org}
} Note that IPOPT is a general-purpose nonlinear solver designed for
large-scale problems, while the problems at hand are small and have additional
structure. A dedicated solver can leverage this structure for improved
performance~\cite{caron2018capturability}.

We ran a benchmark over randomized states and contact locations for
Problems~\ref{cvx-prob-2d} and \ref{cvx-prob-3d}. Both solvers were fed the
same initial states, and sampling was biased toward viable states using the
proxy distribution $z_\text{crit} \sim \frac12 \zf + \sigma_z {\cal U}([0,
1])$. On an Intel Core i7-6500U CPU @ 2.50 Ghz, computation times were
identical: $1.8 \pm 0.7$ ms and $1.7 \pm 0.6$ ms for the 2D and 3D problem,
respectively (averages and standard deviations over $10000$ control cycles
aggregated over 225 launches from different initial states). These times can be
improved by two orders of magnitude using the dedicated solver introduced in
the extension~\cite{caron2018capturability} of the present work.

\begin{figure}[t]
    \centering
    \includegraphics[width=0.9\columnwidth]{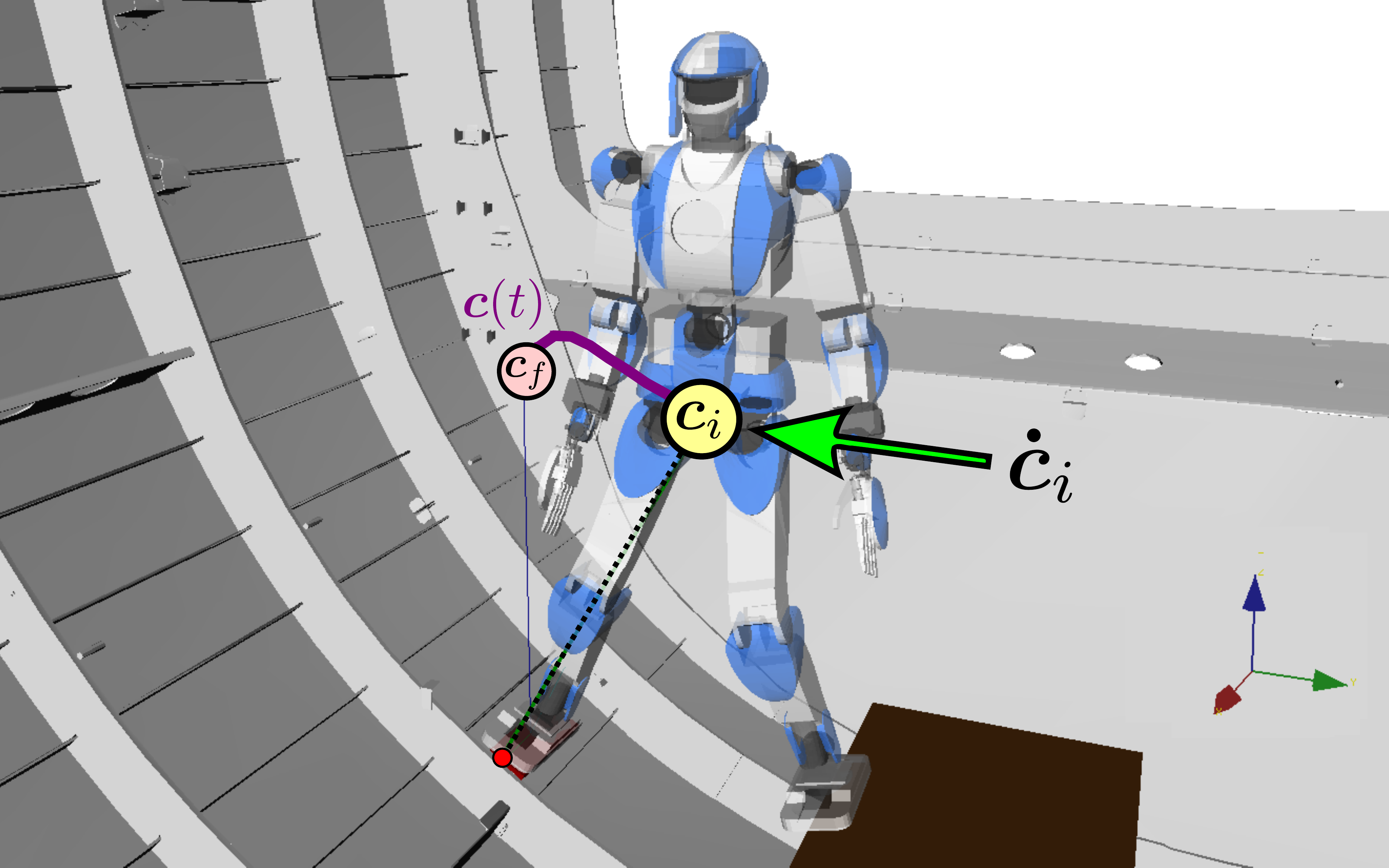}
    \caption{
        \textbf{3D balance control using both CoP and CoM height variations.}
        The robot recovers from an undesired lateral push by stepping onto the
        fuselage of an A350 aircraft under construction. An initial velocity of
        1.4~m\,s$^{-1}$ is absorbed in 1.5~s using both the ankle strategy and
        a 10~cm CoM height variation.
    }
    \label{fig:aircraft}
\end{figure}

We next considered a push recovery scenario for an HRP-4 humanoid evolving in a
3D model of an A350 aircraft under construction. Due to \emph{e.g.}~an
unexpected collision, or slippage over a ground obstacle that is only detected
after momentum has built up, the robot is imparted with an initial velocity of
1.4~m\,s$^{-1}$. Although hand contacts would be equally important in such
scenarios, we focus here on the stance leg trajectory. First, a foothold is
chosen on the fuselage as the kinematically reachable location with the lowest
tilting. The velocity in the resulting frame (Figure~\ref{fig:inertial-frame})
consists of roughly 1.3~m\,s$^{-1}$ in the desired direction of motion
$\bfe_x$, 0.2~m\,s$^{-1}$ in the lateral direction $\bfe_y$ and 0.5~m\,s$^{-1}$
along the vertical $\bfe_z$. We confirmed that the robot is able to stop in
1.5~s using using the CoP strategy~\eqref{eq:f-and-F} with $k=2$ and a 10~cm
CoM height variation. The scenario is depicted in Figure~\ref{fig:aircraft} and
in the accompanying video.

\section{Concluding note}

We saw how the boundedness condition can lead to an alternative optimization
for 3D control of the inverted pendulum model.
See~\cite{caron2018capturability} for an extension of this approach to 3D
bipedal walking, addressing follow-up questions such as contact switches and
the efficient resolution of the underlying numerical optimization.

\section*{Acknowledgment}

We warmly thank Adrien Escande, Leonardo Lanari, Twan Koolen, Michael Posa,
Patrick Wensing and Pierre-Brice Wieber for their helpful suggestions,
corrections or comments on this work.

\bibliographystyle{IEEEtran}
\bibliography{refs}

\appendix

\subsection{Convexity of Equation~\texorpdfstring{\eqref{eq:conv-obj}}{(30)}}
\label{convexity-proof}

Let us first consider the function $g(x, y) = \frac{1}{\sqrt{x} + \sqrt{y}}$.
The trace and determinant of its Hessian $\nabla^2 g$ are:\footnote{
    To avoid painstaking calculations such as this one, we used the
    online computational-knowledge engine {Wolfram$\vert$Alpha} provided by
    Wolfram Research, Inc.: \url{https://www.wolframalpha.com/}
}
\begin{align}
    \tr(\nabla^2 g) & = \frac{x^2 + 3 (x + y) \sqrt{x y} + y^2}{4
    (xy)^{3/2} (\sqrt{x} + \sqrt{y})^3}
    \\
    \det(\nabla^2 g) & = \frac{3}{16 (x y)^{3/2} (\sqrt{x} + \sqrt{y})^4}
\end{align}
Both are strictly positive quantities over the domain $x, y>0$, therefore
$\nabla^2 g$ is positive definite and $g$ is convex. Consider now the function $G(\bfvarphi)$ defined by Equation~\eqref{eq:conv-obj}
over the vector $\bfvarphi$ of positive values $\varphi_1, \ldots, \varphi_N$:
$G(\bfvarphi) = \sum_{j=0}^{N-1} \Delta_j g(\varphi_{j+1}, \varphi_j)$. For any $t
\in [0, 1]$, $ G(t \bfvarphi + (1 - t) \bfvarphi')$
\begin{align}
    & = \sum_{j=0}^{N-1} \Delta_j g(t \varphi_{j+1} + (1 - t) \varphi'_{j+1}, t
    \varphi_{j} + (1 - t) \varphi'_{j}) \nonumber \\
    & \leq
    \sum_{j=0}^{N-1} \Delta_j \left[t g(\varphi_{j+1}, \varphi_j) + (1 - t) g(\varphi'_{j+1}, \varphi'_j)\right] \\
    & = t G(\bfvarphi) + (1 - t) G(\bfvarphi')
\end{align}
Which establishes that $G$ is convex. \QED

\subsection{Computing time trajectories}
\label{computing-time-traj}

The piecewise constant values of $\lambda(s)$ are directly given by $\lambda_j
= (\varphi_{j+1} - \varphi_j) / \Delta_j$. Computing the time trajectory $\lambda(t)$
is then equivalent to finding the switch times $t_j$ such that $s(t_j) = s_j$.
Solving the equation of motion~\eqref{eq:ipm} of the IPM with constant
$\lambda$, one can establish the recurrence relation:
\begin{equation}
    t_j = t_{j+1} + \frac{1}{\sqrt{\lambda_j}} \log \left(
    \frac{\sqrt{\varphi_{j+1}} + \sqrt{\lambda_j} s_{j+1}}{\sqrt{\varphi_j} +
    \sqrt{\lambda_j} s_j}
    \right)
\end{equation}
The same relation can be applied to find the map $s(t)$, giving $\omega(t) =
\omega(s(t)) = [\varphi_j + \lambda_j (s(t)^2 - s_j^2)]^{1/2} s(t)^{-1}$ for $t
\in [t_j, t_{j+1}]$ by Equation~\eqref{eq:varphi-s}. Alternatively, one can solve
the differential equation~\eqref{eq:omegad} from $t_j$ to get directly:
\begin{equation}
    \omega(t) = \sqrt{\lambda_j} \frac{1 - \upsilon_j \tanh(\sqrt{\lambda_j} (t -
    t_j))}{\upsilon_j - \tanh(\sqrt{\lambda_j} (t - t_j))}
\end{equation}
where $\upsilon_j = \lambda_j / \omega(s_j) = \lambda_j s_j / \varphi_j$.

\end{document}